\documentclass[lettersize,journal]{IEEEtran}
\usepackage{amsmath,amsfonts}
\usepackage{array}
\usepackage[caption=false,font=normalsize,labelfont=sf,textfont=sf]{subfig}
\usepackage{textcomp}
\usepackage{stfloats}
\usepackage{url}
\usepackage{verbatim}
\usepackage{graphicx}
\usepackage{cite}

\usepackage[colorlinks,linkcolor=red]{hyperref}
\usepackage{amssymb}
\usepackage{multirow}
\usepackage{booktabs}
\usepackage{xcolor}
\usepackage{pifont}
\usepackage{color, colortbl}
\definecolor{LightCyan}{rgb}{0.88,1,1}
\usepackage{algorithm,algpseudocode}

\begin{document}

\title{STAL3D: Unsupervised Domain Adaptation for 3D Object Detection via Collaborating Self-Training and Adversarial Learning}

\author{Yanan~Zhang,
        Chao~Zhou,
        and Di~Huang,~\IEEEmembership{Senior~Member,~IEEE}
        % <-this % stops a space
\thanks{This work is partly supported by the National Natural Science Foundation of China (62022011), the Research Program of State Key Laboratory of Software Development Environment, and the Fundamental Research Funds for the
Central Universities. \emph{(Corresponding author: Di Huang.)}

Y. Zhang, C. Zhou and D. Huang are with the State Key Laboratory of Software Development Environment, School of Computer Science and Engineering, Beihang University, Beijing 100191, China. D. Huang is also with the
Hangzhou Innovation Institute, Beihang University, Hangzhou 310051, China (e-mail: zhangyanan@buaa.edu.cn; zhouchaobeing@buaa.edu.cn; dhuang@buaa.edu.cn).}}% <-this % stops a space
%\thanks{Manuscript received April 19, 2021; revised August 16, 2021.}}

\markboth{IEEE Transactions on Intelligent Vehicles,~Vol.~XX, No.~X, May~2024}%
{Zhang \MakeLowercase{\textit{et al.}}: STAL3D: Unsupervised Domain Adaptation for 3D Object Detection via Collaborating Self-Training and Adversarial Learning}


\maketitle

\begin{abstract}
Existing 3D object detection suffers from expensive annotation costs and poor transferability to unknown data due to the domain gap, Unsupervised Domain Adaptation (UDA) aims to generalize detection models trained in labeled source domains to perform robustly on unexplored target domains, providing a promising solution for cross-domain 3D object detection. Although Self-Training (ST) based cross-domain 3D detection methods with the assistance of pseudo-labeling techniques have achieved remarkable progress, they still face the issue of low-quality pseudo-labels when there are significant domain disparities due to the absence of a process for feature distribution alignment. While Adversarial Learning (AL) based methods can effectively align the feature distributions of the source and target domains, the inability to obtain labels in the target domain forces the adoption of asymmetric optimization losses, resulting in a challenging issue of source domain bias. To overcome these limitations, we propose a novel unsupervised domain adaptation framework for 3D object detection via collaborating ST and AL, dubbed as STAL3D, unleashing the complementary advantages of pseudo labels and feature distribution alignment. Additionally, a Background Suppression Adversarial Learning (BS-AL) module and a Scale Filtering Module (SFM) are designed tailored for 3D cross-domain scenes, effectively alleviating the issues of the large proportion of background interference and source domain size bias. Our STAL3D achieves state-of-the-art performance on multiple cross-domain tasks and even surpasses the Oracle results on Waymo $\rightarrow$ KITTI and Waymo $\rightarrow$ KITTI-rain.
\end{abstract}


\section{Introduction}
\IEEEPARstart{3D}{object} detection is crucial for the perception systems of autonomous driving, aiming to classify and localize objects in the real-world 3D space, providing essential groundwork for higher-level tasks such as trajectory prediction and path planning. Recently, significant progress~\cite{tian2023acf,zhang2022cat,ai2023lidar,cao2023mchformer,wang2023multi,min2023occupancy,yang2023sacinet,zhang2023sa} has been made in this task due to the advancement of deep learning techniques and the emergence of large-scale annotated datasets~\cite{geiger2012we,sun2020scalability,caesar2020nuscenes,lyft2019} for autonomous driving. However, due to the presence of domain shift, when applying models trained on known domains directly to new domains, performance significantly deteriorates, hindering the generalizability and transferability of detectors across different scenes.

To overcome such challenge, Unsupervised Domain Adaptation (UDA) strives to transfer knowledge from a labeled source domain to an unlabeled target domain. While there have been numerous studies on UDA in the field of 2D object detection, these methods are not applicable to sparse, unordered, and irregular point clouds. As a result, 3D UDA methods have not been thoroughly explored yet. The existing cross-domain 3D detection methods are mainly divided into two paradigms, namely the Self-Training (ST) paradigm and the Adversarial Learning (AL) paradigm.

\begin{figure}[t]
\centering
\includegraphics[width=0.99\columnwidth]{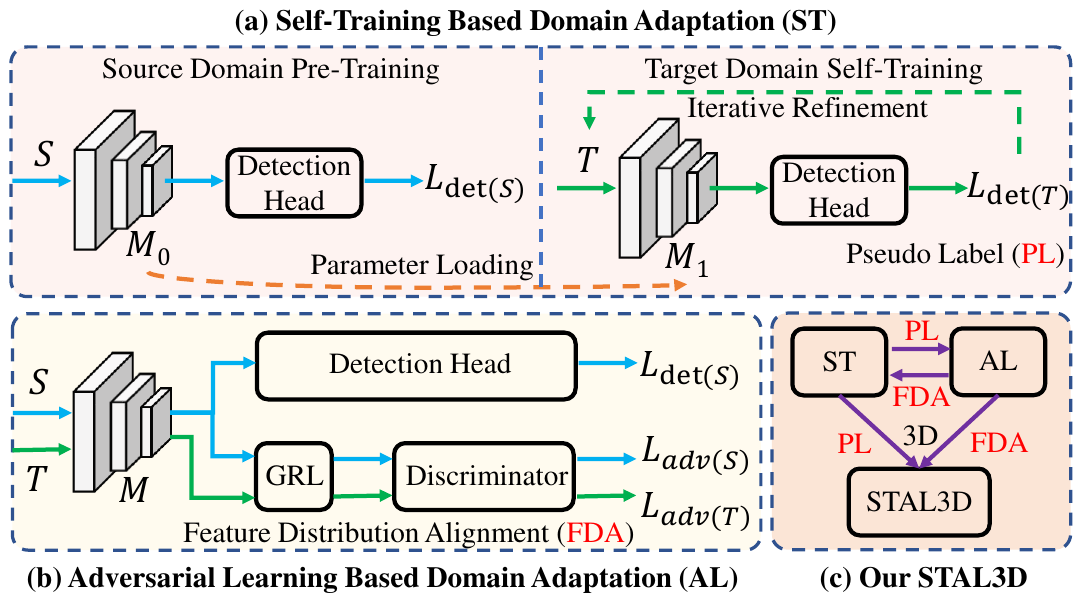}
\caption{Domain Adaptation Paradigms for 3D object detction. (a) Self-Training Based Domain Adaptation (ST); (b) Adversarial Learning Based Domain Adaptation (AL); (c) Collaborating Self-Training and Adversarial Learning (Our STAL3D).}
\label{first}
\end{figure}
As shown in Fig.~\ref{first}(a), the ST paradigm~\cite{wang2020train,yang2021st3d,yang2022st3d++,hu2023density,wei2022lidar} is the simplest and most straightforward scheme to narrowing the domain gap in domain adaptation, which primarily consists of a source domain pre-training stage and a target domain self-training stage. During the initial stage, the model is trained under supervision using labeled data from the source domain. Subsequently, the parameters of the trained model are loaded to generate pseudo-labels for the target domain, and the model is iteratively updated and trained using these pseudo-labels. To accommodate the peculiarities of 3D data, certain methods take into account the disparities in the authentic 3D physical size distribution of objects, proposing size distribution regularization~\cite{wang2020train} and size data augmentation techniques~\cite{yang2021st3d,yang2022st3d++}. Alternatively, some approaches consider the differences in LiDAR sensor scanning beam patterns and devise beam resampling~\cite{hu2023density} or LiDAR distillation~\cite{wei2022lidar} strategies. However, these methods only consider domain differences along a single dimension, overlooking the significant domain differences caused by various factors such as weather, road conditions, sensor types, etc.
Due to the inherent lack of feature distribution alignment process in ST, when there are significant domain disparities, they will generate low-quality pseudo-labels. These low-quality pseudo-labels have a detrimental effect on model optimization, leading to error accumulation during the iterative process of pseudo-label generation and model updates, ultimately resulting in performance degradation.

As shown in Fig.~\ref{first}(b), the AL paradigm~\cite{wang2019range,zhang2021srdan,yihan2021learning} comprises a shared feature extractor and a domain discriminator, which simultaneously take annotated data from the source domain and unlabeled data from the target domain as input.
Through Gradient Reversal Layer (GRL) and domain discriminator, the source and target domains minimize the distribution discrepancy under the effect of adversarial loss. Additionally, the labeled source domain is also optimized through detection loss. Inspired by learning domain-invariant feature representations in 2D object detection, some methods have explored the application of adversarial learning to 3D domain adaptation from perspectives such as range perception~\cite{wang2019range}, scale perception~\cite{zhang2021srdan}, and category perception~\cite{yihan2021learning}. However, due to the lack of pseudo-labels, the target domain only utilizes adversarial loss $L_{adv}$ for optimization, while the source domain can optimize using both detection loss $L_{det}$ and adversarial loss $L_{adv}$ simultaneously. The asymmetric optimization losses hinder features from aligning to a balanced position between two domains, leading to a source-bias issue that significantly compromises the detectors' generalization capability in the target domain.

Through the above analysis, we can observe that: (1) The ST paradigm excels in providing pseudo-label supervision signals for unlabeled target domains, yet its inherent limitation lies in the lack of feature distribution alignment, posing challenges in simultaneously adapting to multiple domain disparities, especially in the presence of significant domain disparities, leading to the generation of low-quality pseudo-labels; (2) The advantage of the AL paradigm lies in its capability to address disparities from multiple domains and significant domain gaps through feature distribution alignment. However, its drawback is the lack of supervision signals from the target domain, forcing the formation of asymmetric optimization losses, which can lead to the issue of source domain bias.

Motivated by the strong complementarity between the two paradigms, as shown in Fig.~\ref{first}(c), we propose a novel unsupervised domain adaptation framework for 3D object detection via collaborating ST and AL, unleashing the potential advantages of pseudo-labels and feature distribution alignment. From the perspective of ST to AL, our ST approach can generate pseudo-labels for unlabeled target domain data, which then participate in the training process of AL. By obtaining additional pseudo-label supervision signals, AL forms symmetric optimization losses, namely adversarial loss and detection loss for both the source and target domains, which can effectively address the issue of source domain dominance previously caused by asymmetric gradient optimization. From the perspective of AL to ST, our AL approach utilizes gradient reversal layer and domain discriminator to incorporate additional feature distribution alignment constraints into the feature extraction network of ST, thereby forming domain-invariant features. Relying on domain-invariant feature representation, ST can generate higher-quality pseudo-labels even when confronted with multiple domain disparities or significant domain gaps, thus effectively alleviating the accumulation of errors during the iterative process.

Additionally, the task of 3D cross-domain object detection exhibits some distinct characteristics compared to traditional 2D tasks: (1) In 3D scenes, the proportion of background is significantly larger than that of foreground, leading to potential background interference; (2) 3D detection reflects the real size of the object in the physical world, but the size distribution of the same category in different domains varies greatly, resulting in a unique source domain size bias problem. Regarding the above two issues, we design a Background Suppression Adversarial Learning (BS-AL) module and a Scale Filtering Module (SFM) to alleviate the issues of the large proportion of background interference and source domain size bias, respectively.

In summary, our contributions are as follows:

\begin{itemize}
  \item We point out the strong complementarity between ST and AL and propose a novel collaborative STAL3D framework for cross-domain 3D object detection, unleashing the potential advantages of pseudo-labels and feature distribution alignment.
  \item We design a Background Suppression Adversarial Learning (BS-AL) module and a Scale Filtering Module (SFM) tailored for 3D cross-domain scenes, effectively alleviating the issues of the large proportion of background interference and source domain size bias.
  \item We conduct extensive experiments on multiple datasets for three categories. The proposed STAL3D consistently outperforms the strong baseline by large margins, highlighting its effectiveness.
\end{itemize}

\section{Related Work}

\begin{figure*}[t]
\centering
\includegraphics[width=0.99\textwidth]{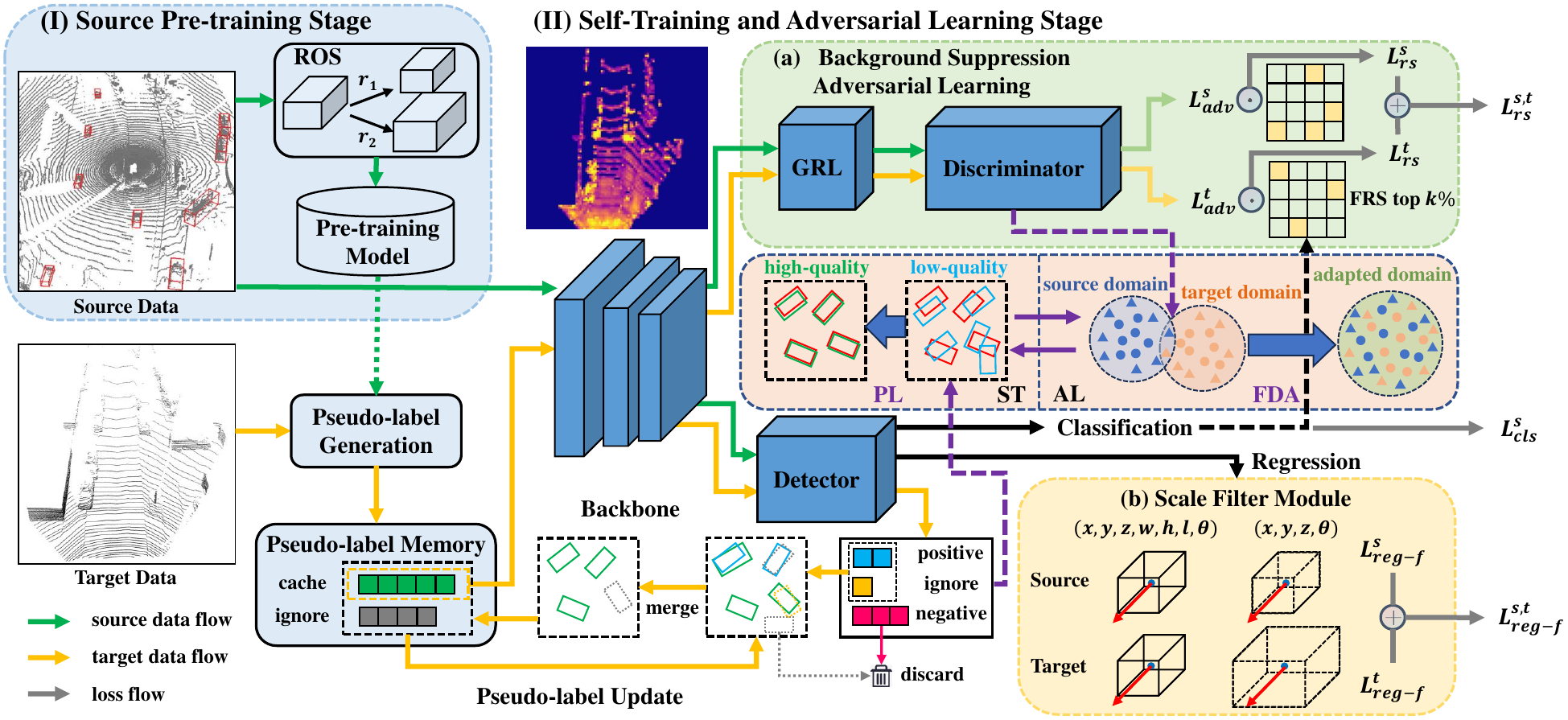}
\caption{Overview of the proposed STAL3D framework. The training process includes two stages: (I) Source Pre-training Stage; (II) Self-Training and Adversarial Learning Stage. In stage (I), Random Object Scaling (ROS) data augmentation is used for pre-training on the source domain dataset to obtain model initialization. In stage (II), the model is training via collaborating Self-Training with our Scale Filter Module (SFM) and Background Suppression Adversarial Learning (BS-AL), unleashing the complementary advantages of pseudo-labels and feature distribution alignment.}
\label{fig2_framework}
\end{figure*}

\subsection{LiDAR-based 3D Object Detection}
LiDAR-based 3D object detection techniques can be broadly classified into three categories: Point-based methods, Voxel-based methods and Point-Voxel based methods. Point-based methods~\cite{qi2018frustum, shi2019pointrcnn,zhang2021pc,lu2023improving}, employing PointNet~\cite{qi2017pointnet, qi2017pointnet++} as the backbone network, directly extract geometric features from raw point clouds to accomplish detection tasks. Recent advancements~\cite{yang20203dssd, chen2022sasa, zhang2022not, he2023sa} in this area have focused on enhancing performance through the design of more effective point sampling strategies. However, these approaches often necessitate time-consuming point sampling and neighbor search operations. Voxel-based methods~\cite{zhou2018voxelnet, yan2018second, lang2019pointpillars, yin2021center} typically partition point clouds into regular grid structures and employ 3D convolutional backbones to extract features. Recent advancements~\cite{mao2021voxel, he2022voxel, fan2022embracing, zhou2023octr} have also explored leveraging Transformer architectures to enhance the representation capabilities of voxel features by capturing long-range dependency relationships. While computationally efficient, voxelization inevitably introduces quantization loss. Point-voxel based methods~\cite{yang2019std,shi2020pv,he2020structure}, on the other hand, endeavor to amalgamate the advantages of both voxel-based and point-based approaches. However, existing approaches overlook domain discrepancies across different 3D scenes, making them almost inapplicable to unseen environments. In this paper, we investigate domain adaptation in 3D object detection, which effectively enhances the domain generalization capabilities of leading 3D detectors.

\subsection{Cross-domain 3D Object Detection}
Existing cross-domain 3D detection methods primarily fall into two paradigms: Self-Training and Adversarial Learning. The Self-Training (ST) paradigm mainly focuses on leveraging annotated data from the source domain to pre-train robust initial models and optimize the quality of generated pseudo-labels. Zoph et al.~\cite{zoph2020rethinking} leverage techniques like data distillation and self-training with data augmentation to mitigate confirmation bias and improve pseudo-label quality. Wang et al.~\cite{wang2020train} first claim that the size distribution of objects in different datasets is a key factor influencing the model's domain adaptation performance and propose a simple strategy to correct sizes using statistical information. ST3D~\cite{yang2021st3d} introduces a data augmentation technique to enhance the robustness of pre-trained models to target sizes and proposes a quality-aware triplet memory bank to refine pseudo-labels. ST3D++~\cite{yang2022st3d++} conducts further analysis on pseudo-label noise and proposes a pseudo-label denoised self-training pipeline from pseudo-label generation to model optimization. DTS~\cite{hu2023density} proposes a density-insensitive cross-domain approach to mitigate the impact of domain gap resulting from varying density distributions. AVP~\cite{li2023adaptation} explicitly leverages cross-domain relationships to efficiently generate high-quality samples, thereby mitigating domain shifts. The Adversarial Learning (AL) paradigm strives to reduce domain gap by better aligning the feature distributions of the source and target domains. Wang et al.~\cite{wang2019range} for the first time attempt to study adaptation for 3D object detection in point clouds, which combine fine-grained local adaptation and adversarial global adaptation to improve LiDAR-based far-range object detection. SRDAN~\cite{zhang2021srdan} additionally introduces domain alignment techniques in both scale and range, exploiting the geometric characteristics of point clouds for aligning feature distributions. 3D-CoCo~\cite{yihan2021learning} utilizes a contrastive learning mechanism to minimize feature distances within the same category across various domains while maximizing feature distances between different categories, facilitating the alignment of feature distributions. However, existing methods have not fully explored the advantages, disadvantages, and complementary effects of ST and AL paradigms. In this paper, we conduct an in-depth analysis of the inherent synergy between ST and AL and propose a novel framework to unleash the potential advantages of pseudo-labels and feature distribution alignment.

\section{Method}

\subsection{Framework Overview}
As shown in Fig.~\ref{fig2_framework}, the framework primarily consists of two stages: source domain pre-training stage and self-training and adversarial learning stage. In the source domain pre-training stage, we conduct supervised training using annotated source data to obtain initial parameters. During the self-training and adversarial learning stage, in the target domain, we use pre-trained parameters from stage (I) to generate pseudo-labels. Simultaneously, Background Suppression Adversarial Learning Module (BS-AL) is applied to perform feature distribution alignment for source and target domains. Additionally, Scale Filtering Module (SFM) is applied to  alleviate the issue of source domain size bias.

In the scenario of unsupervised domain adaptation, we are provided with point cloud data from a single labeled source domain $\mathcal{D}_S = \{(P^s_i, L^s_i)\}^{n_s}_{i=1}$ and an unlabeled target domain $\mathcal{D}_T = \{P^t_i\}_{i=1}^{n_t}$, where ${n_s}$ and ${n_t}$ denote the number of samples ($P$ is the point clouds, $L$ is the corresponding label) from the source and target domains, respectively. The 3D box annotation for the $i$-th point cloud in the source domain is represented as $L^s_i= \{b_j\}^{B_i}_{j=1}$, with $b_j = (x, y, z, w, l, h, \theta, c)\in\mathbb{R}^8$, where $B_i$ indicates the total number of labeled boxes in $P^s_i$. Here, $(x, y, z)$ denotes the center location of the box, $(w, l, h)$ represents the box dimensions, $\theta$ signifies the box orientation, and $c \in \{1, ..., C\}$ denotes the object category. The objective of the domain adaptive detection task is to train a model $F$ utilizing $\mathcal{D}_S$ and $\mathcal{D}_T$, with the aim of maximizing performance on $\mathcal{D}_T$. The training process is outlined in the following Algorithm~\ref{algo:our_pipeline}.

\begin{algorithm}[t]
	\small
	\caption{Pipeline of our STAL3D.}
	\renewcommand{\algorithmicensure}{ \textbf{Output:}}
	\label{algo:our_pipeline}
    \begin{algorithmic}[1]
		\Require
		Source domain labeled data $\{(P^s_i, L^s_i)\}^{n_s}_{i=1}$, and target domain unlabeled data $\{P^t_i\}_{i=1}^{n_t}$.
		\Ensure The object detection model $F(\cdot;\theta)$ for target domain.
		
		\State Pre-train the object detector $F$ on source domain data $\{(P^s_i, L^s_i)\}^{n_s}_{i=1}$ with ROS as elaborated in Sec.~\ref{pretraining} and obtain the pre-trained parameters of the model $\theta_{ros}$.
		\State Utilize the current model to generate {raw object proposals} $[B^{t}_i]_k$ for every target domain sample $P_i^t$, where $k$ is the current number of times for pseudo label generation.
		\State Update the memory $[M_i^{t}]_k$ given current pseudo labels $[\hat{L}^{t}_i]_k$ and historical pseudo labels $[M_i^{t}]_{k-1}$ in the memory as described in Sec.~\ref{self-training}. The memory $\{[M^{t}_i]_k\}_{i=1}^{n_t}$ consists of pseudo labels for all unlabeled target domain samples.
		\State Train the model on target domain data $\{P_i^t, [M_i^{t}]_k\}_{i=1}^{n_t} $ and source domain data $\{(P^s_i, L^s_i)\}^{n_s}_{i=1}$ with BS-AL in Sec.~\ref{BS-AL} and SFM in Sec.~\ref{SF_Loss} for several epochs.
		\State Go back to Line 2 until convergence.
	\end{algorithmic}
\end{algorithm}

\subsection{Source Domain Pre-training}
\label{pretraining}
STAL3D starts from training a 3D object detector on labeled source data  $\{(P^s_i, L^s_i)\}^{n_s}_{i=1}$. The pre-trained model learns how to perform 3D detection on source labeled data and is further adopted to initialize object predictions for the target domain unlabeled data.

As shown in Fig.~\ref{fig2_framework}(I), following~\cite{yang2021st3d}, we first use Random Object Scale (ROS) to initially alleviate the domain disparities in size distribution by data augmentation. Let $\{p_{i}\}_{i=1}^{n_{b}}$ denote all the points within the annotated box $B_{b}$, where $p_{i}$ is represented as the coordinates of the point cloud $(p_{i}^{x}, p_{i}^{y}, p_{i}^{z})$. $(c_{b}^{x}, c_{b}^{y}, c_{b}^{z})$ represents the center point of the annotated box $B_{b}$, and $R$ is the rotation matrix. Then,$\{p_ {i}\}_{i=1}^{n_{b}}$ can be expressed in the target's center coordinate system as follows:
\begin{equation}
    \label{eq1}
    (p_{i}^{l}, p_{i}^{w}, p_{i}^{h}) = (p_{i}^{x}-c_{b}^{x}, p_{i}^{y}-c_{b}^{y}, p_{i}^{z}-c_{b}^{z}) \cdot R
\end{equation}
After obtaining the coordinates in the target's center coordinate system, a random scaling factor within a certain range $(r_{l}, r_{w}, r_{h})$ is set. The augmented point cloud can be represented as $p_{i}^{aug}$, as shown in Equation~\ref{eq2}.
This augmentation is applied to the source domain data for pre-training, resulting in the pre-trained parameters of the model $\theta_{ros}$.
\begin{equation}
    \label{eq2}
p_{i}^{aug} = (r_{l}p_{i}^{l}, r_{w}p_{i}^{w}, r_{h}p_{i}^{h}) \cdot R^{T} + (c_{b}^{x}, c_{b}^{y}, c_{b}^{z})
\end{equation}

Our framework consists of two key stages: source domain pre-training and iterative self-training. The purpose of the source domain pre-training stage is to provide a robust initialization model for the iterative self-training stage. Additionally, unlike 2D object detection tasks, 3D object detection reflects the real sizes of objects in 3D physical space, and differences in the distribution of object sizes across datasets can affect the training of the pre-trained model, thereby influencing the initialization effect of iterative self-training. Therefore, we consider using this random scaling strategy to augment the object point cloud during the source domain pre-training phase for improving the training quality of the initial model.

\begin{figure}[t]
\centering
\includegraphics[width=0.99\columnwidth]{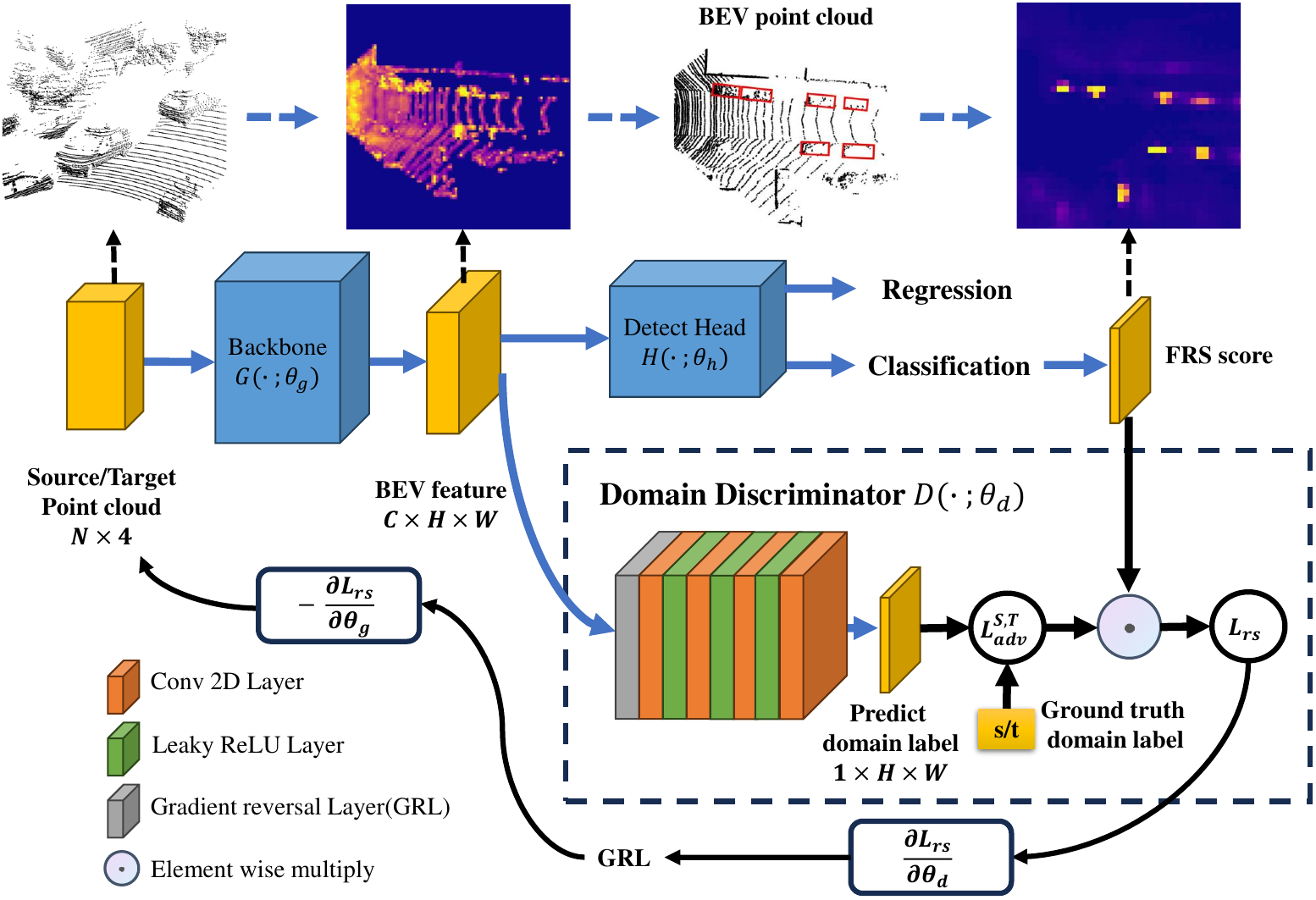}
\caption{Background Suppression Adversarial Learning. A Feature Richness Score (FRS) is used to divide the entire scene into a learning region and a suppression region. Using FRS map to weight the original adversarial training loss, the model focuses on more valuable foreground regions, avoiding noise interference introduced by background.}
\label{al}
\end{figure}

\subsection{Self-Training}
\label{self-training}
With the trained detector, the self-training step is to generate pseudo labels and perform iterative refinement for the unlabeled target data.

As shown in Fig.~\ref{fig2_framework}(II), we introduce the Pseudo-Label Memory Bank Integration module. This module takes the pseudo-labels from the $k$-th stage $\{\hat{L_{i}^{t}}\}_{k}^{n_{l}}$ and the pseudo-labels stored in the memory bank $\{\hat{M_{i}^{t}}\}_{k-1}^{n_{m}}$ as input, and outputs the integrated memory bank pseudo-labels $\{\hat{M_{i}^{t}}\}_{k}$, where ${n_l}$ and ${n_m}$ denote the number of pseudo-labels from the present model and memory bank, respectively. Specifically, we calculates a 3D IoU matrix $A=\{a_{ef}\} \in \mathbb{R}^{n_{m} \times n_{l}}$ between the two sets of labels. For the $e$-th target box in the memory bank, its matched target box is determined as $e^{'}=\text{argmax}_{e}(a_{ef})$. If $a_{ee^{'}} \geq 0.1$, the two boxes are considered a match, and the label with the higher score from $\{\hat{M_{e}^{t}}\}_{k-1}$ and $\{\hat{L_{e^{'}}^{t}}\}_{k}$ is selected as the new pseudo-label and stored in the memory bank. If $a_{ee^{'}} < 0.1$, we sets up an additional buffer in the memory bank, caching these target boxes and maintaining them using a queue.

It is worth noting that since our STAL3D framework introduces feature distribution alignment for ST, it will enable our ST to better cope with large domain  disparities, thereby reducing the accumulation of errors during the iteration process caused by low-quality pseudo labels.

\subsection{Background Suppression Adversarial Learning}
\label{BS-AL}

As shown in Fig.~\ref{al}, for aligning the feature distributions between the source domain and the target domain,
%inspired by Domain Adversarial Neural Networks (DANN)~\cite{ganin2015unsupervised},
we employ adversarial learning for feature distribution alignment. The detector is divided into a backbone network $G$ and a detection head network $H$. For the feature map $F\in \mathbb{R}^{H \times W \times L \times d}$ generated by the backbone network, a domain classifier $D$ is introduced, and a min-max optimization adversarial game loss is constructed. Specifically, the optimization direction for the domain classifier $D$ is to distinguish the source domain from the target domain, minimizing the domain classification loss. Conversely, the optimization direction for the backbone network $G$ is to make it difficult for the domain classifier $D$ to distinguish the source of the features, maximizing the domain classification loss. To achieve end-to-end training, we employs a Gradient Reversal Layer (GRL)~\cite{goodfellow2020generative} to connect $G$ and $D$, where the gradient of the model is inverted when passing through the GRL layer, allowing the optimizer to perform normal optimization. When the model's training converges, the backbone network $G$ can extract domain-invariant feature representations, thereby accomplishing feature distribution alignment.

\begin{figure*}[t]
\centering
\includegraphics[width=0.99\textwidth]{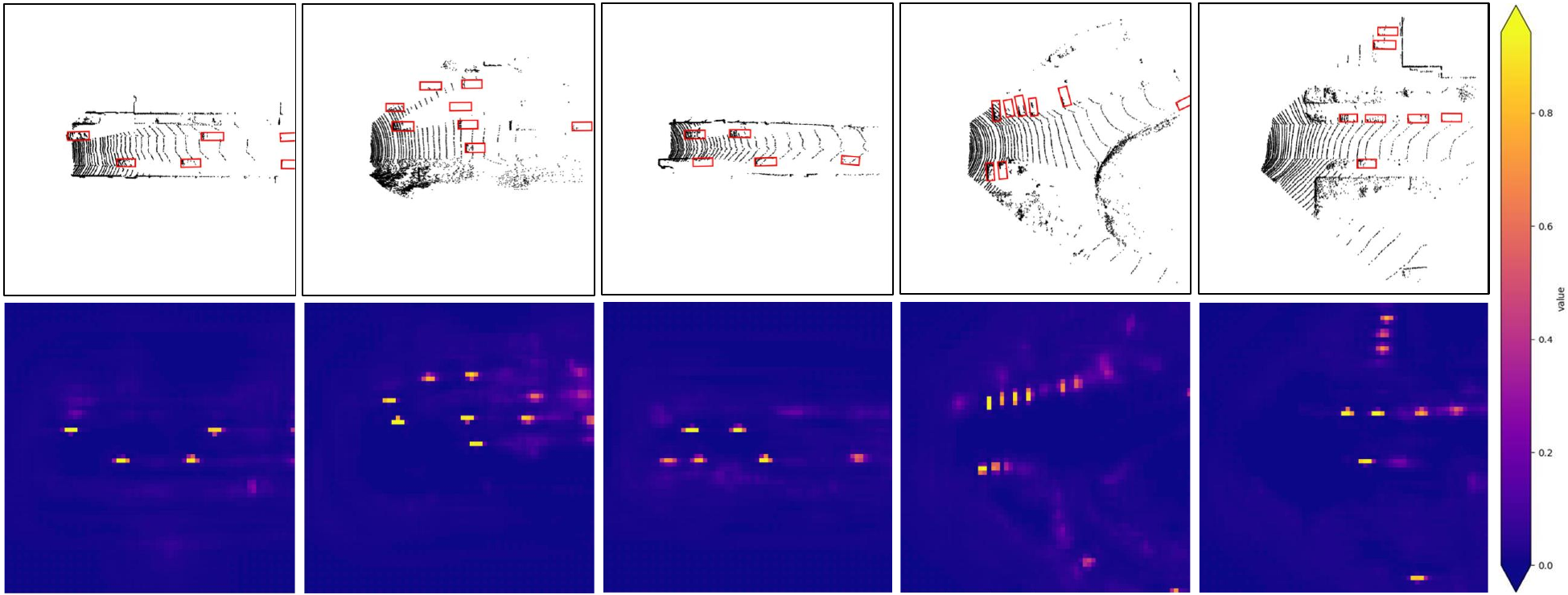}
\caption{Visualization of the Feature Richness Score (FRS) map. The top row represents various point cloud scenes, and the bottom row represents the corresponding FRS map. The brighter the color, the higher the score.}
\label{FRS_map}
\end{figure*}

Let $\mathcal{D}{s}$ and $\mathcal{D}{t}$ represent the source and target domain data respectively, $\theta_{G}$ and $\theta_{D}$ represent the parameters of the backbone network and the domain classifier, and $D(\cdot)^{(h,w,l)}$ represents the probability that the feature at position $(h,w,l)$ in the feature map comes from the source domain. The loss function in this stage can be expressed as:
\begin{equation}
\resizebox{0.99\columnwidth}{!}{
    \label{adv_loss}
$\mathcal{L}_{adv} = \mathop{max}\limits_{\theta_{G}} \mathop{min}\limits_{\theta_{D}} -\mathbb{E}_{x_{s} \sim \mathcal{D}_{s}}\text{log}D(G(x_{s}))-\mathbb{E}_{x_{t} \sim \mathcal{D}_{t}}\text{log}(1-D(G(x_{t})))$
}
\end{equation}
where $x_{s}$, $x_{t}$ and $\mathbb{E}$ respectively stand for inputs from source and target, and the Expectation.

In the context of 3D object detection, the foreground region occupies a small proportion of the entire scene, accounting for only about 5\% \cite{liu2022spatial}. Additionally, in 3D object detection tasks, the foreground region is typically much more important than the background, as it contains much richer semantic information. However, a naive domain classifier does not distinguish between foreground and background, and performing feature distribution alignment with such a classifier can lead to a long-tail problem, causing a decrease in focus on the foreground. Therefore, the Feature Richness Score (FRS) \cite{zhixing2021distilling} of the feature map is utilized as semantic foreground guidance to address this issue. The FRS is then used to apply attention weighting to the adversarial loss based on the foreground region. As shown in Fig.~\ref{FRS_map}, valuable areas can be well distinguished through this approach.

Specifically, let $F_{h,w,l} \in F$ represent the feature vector at position $(h,w,l)$ in the feature map. Let $C$ represent the categories to be detected, and $N_{dir}$ denote the number of predefined directions for anchor boxes. The 3D object detection network uses a 1x1 convolution to predict the confidence of predefined anchor boxes at position $F_{h,w,l}$, resulting in $p_{h,w,l} = \text{conv}1\times1(F_{h,w,l}), p \in \mathbb{R}^{C \times N_{dir}}$. After obtaining the confidence score vector, the feature richness score $S_{h,w,l}$ at position $(h,w,l)$ can be obtained by taking the maximum value of the score vector, as shown in Equation \ref{ch4_rfs}. Here, $\sigma(\cdot)$ represents the sigmoid function.
\begin{equation}
    \label{ch4_rfs}
S_{h,w,l}=\mathop{max} \limits_{i\in[1, C \times N_{dir}]} \sigma(p_{h,w,l}^{i})
\end{equation}

In consideration of the issue that the background occupies a much higher proportion than the foreground and contains less informative content in 3D object detection tasks, the algorithm divides the entire scene into two regions: a learning region (dominated by the foreground) and a suppression region (dominated by the background). The feature richness score is used to guide this division. We considers the voxels with feature richness scores in the top $k\%$ as part of the learning region, while the remaining voxels are considered part of the suppression region. Additionally, the feature richness scores of the suppression region are set to 0. This is expressed in Equation \ref{ch4_region_supp}.
\begin{equation}
    \label{ch4_region_supp}
\hat{S}_{h,w,l} =
\begin{cases}
    S_{h,w,l}, &\ \text{if} \ S_{h,w,l} \ \text{in} \ \text{top} \ k\%
    \\ 0, &\ \text{otherwise.}
\end{cases}
\end{equation}

After obtaining the region partition map $\hat{S} \in \mathbb{R}^{H \times W \times L}$, we proceed to weight the original adversarial training loss using this feature richness score map, resulting in the final adversarial loss with region-based suppression:
\begin{equation}
    \label{ch4_rs_loss}
\mathcal{L}_{rs} = \sum_{h, w, l} \hat{S}_{h,w,l} \cdot \mathcal{L}_{adv}^{h,w,l}
\end{equation}

It is worth noting that due to the introduction of ST process in our STAL3D framework for AL, pseudo-label supervised signals in the target domain can be obtained during the training process, resulting in symmetric optimization loss in AL during the training process, which can effectively alleviate the source domain bias problem.

\begin{figure}[t]
\centering
\includegraphics[width=0.99\columnwidth]{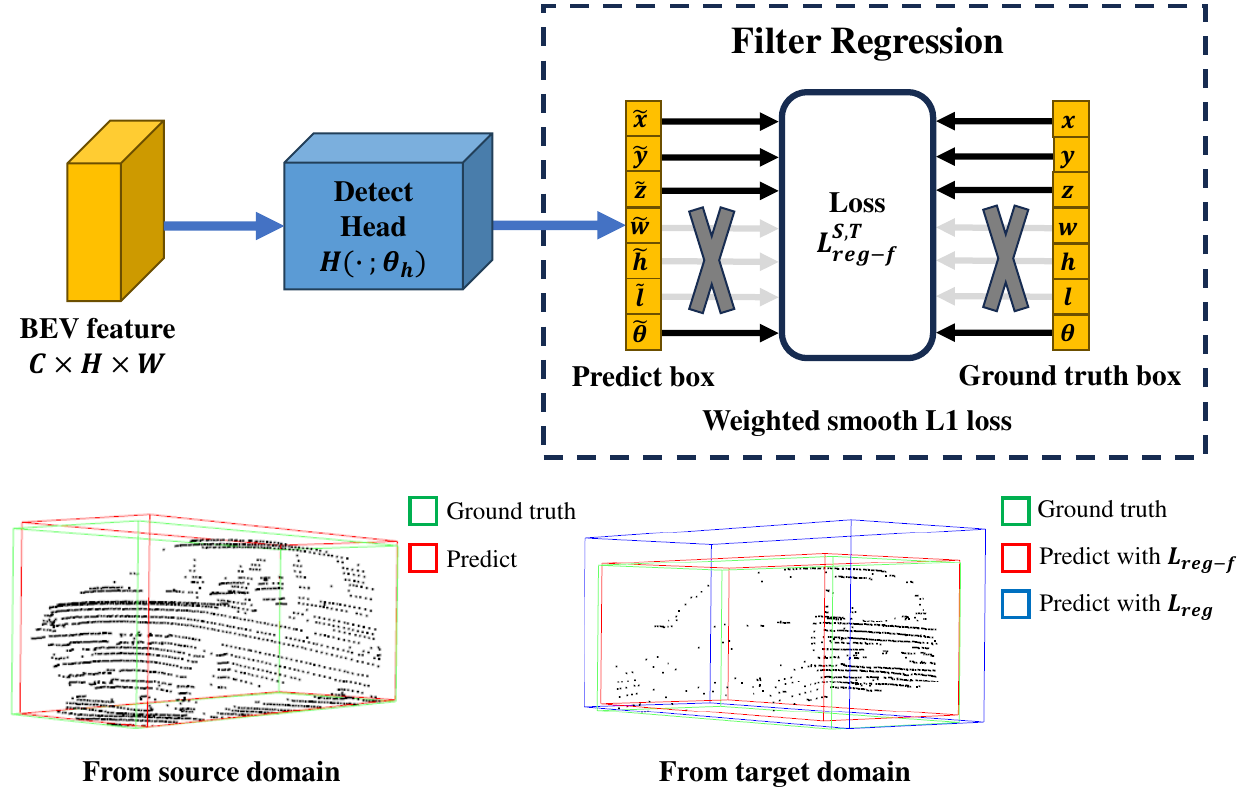}
\caption{Scale Filtering Module. By introducing the loss design of scale regression term filtering, the issue of source domain size bias can be effectively alleviated.}
\label{sfm}
\end{figure}

\begin{figure}[t]
\centering
\includegraphics[width=0.95\columnwidth]{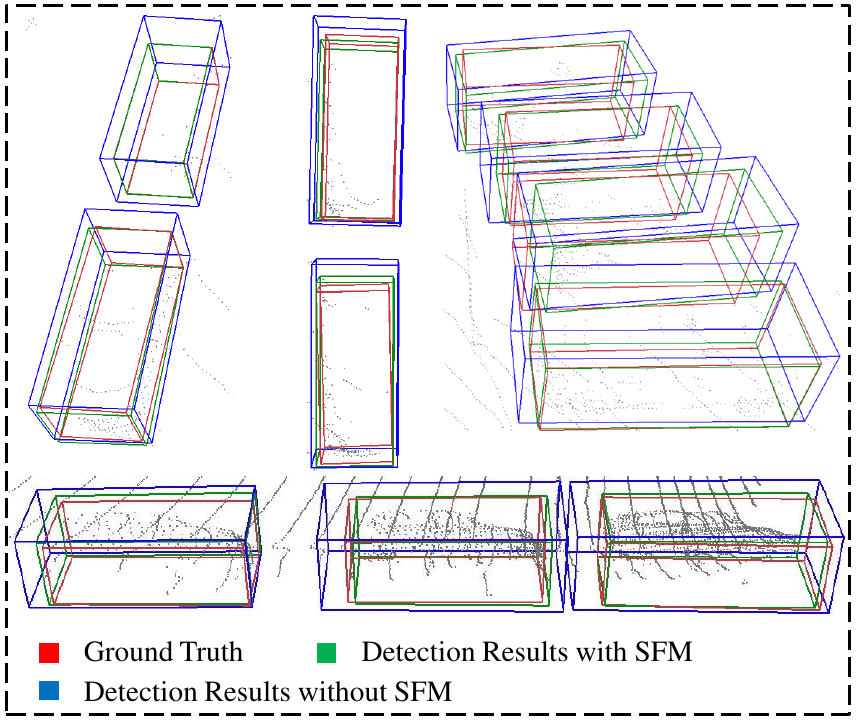}
\caption{Visual comparison of detection results with and without SFM.}
\label{SFM_VIS}
\end{figure}

\subsection{Scale Filtering Module}
\label{SF_Loss}

Compared to pseudo-labels, source domain labels can be regarded as noise-free label, so introducing source domain label supervised signals can theoretically assist in the self-training process. In 3D object detection, the size of objects reflects their dimensions in the real world, with different domains exhibiting significant variations in size distribution. Therefore, directly training the network using source domain label information poses a critical challenge as the model may gradually overfit to the source domain size due to inter-domain size differences throughout the training process, commonly referred to as the issue of source domain size bias.

When dealing with source domain size bias, we begins by addressing the loss design of the 3D object detection model. Taking the single-stage network SECOND~\cite{yan2018second} as an example, its loss design primarily encompasses three components: classification loss, angle classification loss, and regression loss. For the classification loss, the model employs Focal Loss~\cite{lin2017focal} to tackle the foreground-background class imbalance. Due to the strong shape similarity of 3D objects, utilizing target domain pseudo-labels for classification loss might introduce additional noise. Therefore, the model does not compute the classification loss for the pseudo-labels of the target domain data. For angle classification loss, the model employs the cross-entropy loss. As for the regression loss, we employs the Smooth L1 loss to calculate the regression loss. The regression targets, denoted as $(x, y, z, h, w, l, \theta)$, are normalized and encoded as follows:
\begin{equation}
\resizebox{0.99\columnwidth}{!}{
    \label{eq4}
    $\begin{aligned}
x_{t}=\frac{x_{g}-x_{a}}{d_{a}}, y_{t}=\frac{y_{g}-y_{a}}{d_{a}}, & z_{t}=\frac{z_{g}-z_{a}}{h_{a}}, \theta_{t} = \text{sin}(\theta_{g}-\theta_{a}),
\\ h_{t} = \text{log}(\frac{h_{g}}{h_{a}}), w_{t} & = \text{log}(\frac{w_{g}}{w_{a}}), l_{t} = \text{log}(\frac{l_{g}}{l_{a}}),
    \end{aligned}$
    }
\end{equation}
where $x$, $y$, and $z$ represent the center point coordinates, $h$, $w$, and $l$ represent height, width, and length, and $\theta$ represents the rotation angle around the z-axis. Subscripts $t$, $a$, and $g$ correspond to encoded value, anchor value, and ground truth annotation value, respectively. $d_{a}=\sqrt{(l_{a})^{2}+(w_{a})^{2}}$ represents the diagonal length of the anchor box's width and height.

When use of source domain data for predicting object box sizes lacks inter-domain consistency and can lead to severe over-fitting. However, object box localization and angle prediction exhibit inter-domain consistency. Therefore, as shown in Fig.~\ref{sfm}, we filter out regression deviations for the target box sizes $h_{t}, w_{t}, l_{t}$, and only employs $x_{t}, y_{t}, z_{t}, \theta_{t}$ as regression targets. Although this scale filtering design is relatively simple, it has been found to be very effective. As shown in Fig.~\ref{SFM_VIS}, SFM can effectively alleviate the issue of source domain size bias.

Based on this, the overall optimization objective function of the model can be expressed as follows:
\begin{equation}
\resizebox{0.99\columnwidth}{!}{
    \label{ch4_loss}
$\mathcal{L} = \lambda_{1}\mathcal{L}_{FL}^{\mathbf{S}} + \lambda_{2}\mathcal{L}_{reg-f}^{\mathbf{S,T}} + \lambda_{3}\mathcal{L}_{IoU}^{\mathbf{S,T}} + \lambda_{4}\mathcal{L}_{cls-dir}^{\mathbf{S,T}} + \lambda_{5}\mathcal{L}_{rs}^{\mathbf{S,T}}$
}
\end{equation}
where $\mathcal{L}_{FL}$, $\mathcal{L}_{reg-f}$, $\mathcal{L}_{IoU}$, $\mathcal{L}_{cls-dir}$, and $\mathcal{L}_{rs}$ represent the Focal Loss classification loss, filtered regression loss, IoU prediction loss, box orientation classification loss, and region-based adversarial loss with background suppression, respectively. The superscripts $\mathbf{S}$ and $\mathbf{T}$ denote the source and target domains, respectively.

\begin{table*}[t]
\renewcommand\arraystretch{1.15}
    \centering
    \caption{ Experiment results of five adaptation tasks are presented, with the reported average precision (AP) for bird’s-eye view ($\text{AP}_{\text{BEV}}$) / 3D ($\text{AP}_{\text{3D}}$) of car, pedestrian, and cyclist with IoU threshold set to 0.7, 0.5, and 0.5 respectively. When the KITTI is the target domain, we report the AP at \textbf{Moderate} difficulty. The last column shows the mean AP for all classes. We indicate the best adaptation result by \textbf{bold} and we highlight the row representing our method.}
    \resizebox{0.85\textwidth}{!}{
    %\begin{small}
    %\setlength{\tabcolsep}{5.1mm}{
        \begin{tabular}{c|c|c|c|c|c}
            \bottomrule[1pt]
            Task & Method  & Car & Pedestrian & Cyclist & Mean AP \\
            \hline
            \multirow{6}{*}{Waymo $\rightarrow$ KITTI} & Source Only  & 67.64 / 27.48 & 46.29 / 43.13 & 48.61 / 43.84 & 54.18 / 38.15 \\

            & SN~\cite{wang2020train} & 78.96 / 59.20 & 53.72 / 50.44 & 44.61 / 41.43 & 59.10 / 50.36 \\
            % \cline{2-6}
            & ST3D~\cite{yang2021st3d} & 82.19 / 61.83 & 52.92 / 48.33 & 53.73 / 46.09 & 62.95 / 52.08 \\
            & ST3D++~\cite{yang2022st3d++} & 80.78 / 65.64 & 57.13 / 53.87 & 57.23 / 53.43 & 65.05 / 57.65 \\

            & \textbf{STAL3D} \cellcolor{LightCyan} &  \textbf{82.26}\cellcolor{LightCyan} / \textbf{69.78}\cellcolor{LightCyan} & \textbf{59.85}\cellcolor{LightCyan} / \textbf{56.92}\cellcolor{LightCyan} & \textbf{61.33}\cellcolor{LightCyan} / \textbf{57.18}\cellcolor{LightCyan} & \textbf{67.81}\cellcolor{LightCyan} / \textbf{61.29}\cellcolor{LightCyan}  \\
            \cline{2-6}

            & Oracle &  83.29 / 73.45 & 46.64 / 41.33 & 62.92 / 60.32 & 64.28 / 58.37  \\

            \toprule[0.8pt]
            \bottomrule[0.8pt]

            \multirow{6}{*}{Waymo $\rightarrow$ Lyft} & Source Only  & 72.92 / 54.34 & 37.87 / 33.40 & 33.47 / 28.90 & 48.09 / 38.88 \\
            & SN~\cite{wang2020train} & 72.33 / 54.34 & 39.07 / 33.59 & 30.21 / 23.44 & 47.20 / 37.12 \\
            & ST3D~\cite{yang2021st3d} & 76.32 / {59.24} & 36.50 / 32.51 & 35.06 / 30.27 & 49.29 / 40.67 \\
            & ST3D++~\cite{yang2022st3d++} & 79.61 / 59.93 & 40.17 / 35.47 & 37.89 / 34.49 & 52.56 / 43.30 \\

            & \textbf{STAL3D}\cellcolor{LightCyan} & \textbf{81.17}\cellcolor{LightCyan} / \textbf{60.99}\cellcolor{LightCyan} & \textbf{41.75}\cellcolor{LightCyan} / \textbf{37.06}\cellcolor{LightCyan} & \textbf{39.46}\cellcolor{LightCyan} / \textbf{36.12}\cellcolor{LightCyan} &  \textbf{54.13}\cellcolor{LightCyan} / \textbf{44.72}\cellcolor{LightCyan} \\
            \cline{2-6}
            & Oracle &  84.47 / 68.78 & 47.92 / 39.17 & 43.74 / 39.24 & 58.71 / 49.06 \\
            \toprule[0.8pt]
            \bottomrule[0.8pt]

            \multirow{6}{*}{Waymo $\rightarrow$ nuScenes} & Source Only  & 32.91 / 17.24 & 7.32 / 5.01 & 3.50 / 2.68 & 14.58 / 8.31 \\
            & SN~\cite{wang2020train} & 33.23 / 18.57 & 7.29 / 5.08 & 2.48 / 1.80 & 14.33 / 8.48 \\
            & ST3D~\cite{yang2021st3d} & 35.92 / 20.19 & 5.75 / 5.11 & 4.70 / 3.35 & 15.46 / 9.55 \\
            & ST3D++~\cite{yang2022st3d++} & 35.73 / 20.90 & 12.19 / 8.91 & 5.79 / 4.84 & 17.90 / 11.55 \\

            & \textbf{STAL3D}\cellcolor{LightCyan} & \textbf{36.84}\cellcolor{LightCyan} / \textbf{22.13}\cellcolor{LightCyan} & \textbf{15.27}\cellcolor{LightCyan} / \textbf{12.41}\cellcolor{LightCyan} & \textbf{6.37}\cellcolor{LightCyan} / \textbf{5.42}\cellcolor{LightCyan} &  \textbf{19.49}\cellcolor{LightCyan} / \textbf{13.32}\cellcolor{LightCyan} \\
            \cline{2-6}
            & Oracle &  51.88 / 34.87 & 25.24 / 18.92 & 15.06 / 11.73 & 30.73 / 21.84 \\
            \toprule[0.8pt]
            \bottomrule[0.8pt]

            \multirow{6}{*}{nuScenes $\rightarrow$ KITTI}

            & Source Only & 51.84 / 17.92 & 39.95 / 34.57 & 17.70 / 11.08 & 36.50 / 21.19 \\

            & SN~\cite{wang2020train} & 40.03 / 21.23 & 38.91 / 34.36 & 11.11 / 5.67 & 30.02 / 20.42 \\

            & ST3D~\cite{yang2021st3d} & 75.94 / 54.13 & 44.00 / 42.60 & 29.58 / 21.21 & 49.84 / 39.31 \\

            & ST3D++~\cite{yang2022st3d++} & \textbf{80.52} / 62.37 & 47.20 / 43.96 & 30.87 / 23.93 & 52.86 / 43.42 \\

            & \textbf{STAL3D} \cellcolor{LightCyan} & 78.42\cellcolor{LightCyan} / \textbf{65.03}\cellcolor{LightCyan} & \textbf{48.22}\cellcolor{LightCyan} / \textbf{45.73}\cellcolor{LightCyan} & \textbf{32.19}\cellcolor{LightCyan} / \textbf{25.31}\cellcolor{LightCyan} & \textbf{52.94}\cellcolor{LightCyan} / \textbf{45.36}\cellcolor{LightCyan}  \\
            \cline{2-6}
            & Oracle & 83.29 / 73.45 & 46.64 / 41.33 & 62.92 / 60.32 & 64.28 / 58.37 \\
            \toprule[0.8pt]
            \bottomrule[0.8pt]

            \multirow{6}{*}{Lyft $\rightarrow$ KITTI}

            & Source Only & 69.02 / 45.14 & 45.08 / 41.95  & 49.33 / 44.55 & 54.48 / 43.88 \\

            & SN~\cite{wang2020train} & 77.34 / 58.16 & 50.13 / 46.44  & 47.36 / 41.49 & 58.28 / 48.70 \\

            & ST3D~\cite{yang2021st3d} & 79.88 / 64.35 & 50.23 / 47.17  & 54.08 / 46.55 & 61.40 / 52.69 \\

            & ST3D++~\cite{yang2022st3d++} & 80.17 / 66.03  &  54.20 / 51.04  &  57.19 / 53.25  & 63.85 / 56.77\\

            & \textbf{STAL3D} \cellcolor{LightCyan} & \textbf{81.79}\cellcolor{LightCyan} / \textbf{69.22}\cellcolor{LightCyan} & \textbf{55.31}\cellcolor{LightCyan} / \textbf{52.38}\cellcolor{LightCyan} & \textbf{60.54}\cellcolor{LightCyan} / \textbf{56.46}\cellcolor{LightCyan} & \textbf{65.88}\cellcolor{LightCyan} / \textbf{59.35}\cellcolor{LightCyan}  \\
            \cline{2-6}
            & Oracle & 83.29 / 73.45 & 46.64 / 41.33 & 62.92 / 60.32 & 64.28 / 58.37 \\
            \toprule[1pt]
        \end{tabular}

    %}
    %\end{small}
    }
    \label{tab1_main}
\end{table*}

\section{Experiments}

\subsection{Experimental Setup}

\textbf{Datasets.} Our experiments are conducted on four extensively used LiDAR 3D object detection datasets: KITTI~\cite{geiger2012we}, Waymo~\cite{sun2020scalability}, nuScenes~\cite{caesar2020nuscenes}, and Lyft~\cite{lyft2019}. Accordingly, we assess domain adaptive 3D object detection models across the following five adaptation tasks: Waymo $\rightarrow$ KITTI, Waymo $\rightarrow$ Lyft, Waymo $\rightarrow$ nuScenes, nuScenes $\rightarrow$ KITTI and Lyft $\rightarrow$ KITTI.

We tackle the domain shift introduced by adverse weather conditions by simulating rain on the KITTI dataset using the physics-based lidar weather simulation algorithm proposed in~\cite{kilic2021lidar}. By sampling from a range of rain rates spanning from $0 mm/hr$ to $100 mm/hr$ to simulate realistic adverse weather conditions, each sample is augmented with artifacts typical of lidar data captured in rainy weather. Rainy conditions can induce significant domain disparities, resulting in a notable degradation in the quality of point cloud data for vehicles. So, we add two additional domain shift tasks, namely Waymo $\rightarrow$ KITTI-rain and Lyft $\rightarrow$ KITTI-rain.

\textbf{Comparison Methods.} $(i)$ \textbf{Source Only}: Directly evaluates the source domain pre-trained model on the target domain; $(ii)$ \textbf{SN}~\cite{wang2020train}: Pioneering weakly-supervised domain adaptation method for 3D object detection, incorporating statistical object size information from the target domain; $(iii)$ \textbf{ST3D}~\cite{yang2021st3d} and \textbf{ST3D++}~\cite{yang2022st3d++}: State-of-the-art methods based on self-training; $(iv)$ \textbf{Oracle}: Fully supervised model trained exclusively on the target domain.

\begin{table*}[t]
\renewcommand\arraystretch{1.5}
    \centering
    \caption{Adaptation results on Waymo $\rightarrow$ KITTI-rain and Lyft $\rightarrow$ KITTI-rain.}
    \resizebox{0.99\textwidth}{!}{
    \begin{tabular}{c|c|cccc|cccc|cccc}\toprule[1pt]
    {\multirow{2}*{Task}} & {\multirow{2}*{Method}} &\multicolumn{4}{c|}{Car} &\multicolumn{4}{c|}{Pedestrian} &\multicolumn{4}{c}{Cyclist}\\
    & & $\text{mAP}_{BEV}$ & closed gap & $\text{mAP}_{3D}$ & closed gap
    & $\text{mAP}_{BEV}$ & closed gap & $\text{mAP}_{3D}$ & closed gap
    & $\text{mAP}_{BEV}$ & closed gap & $\text{mAP}_{3D}$ & closed gap \\ \bottomrule
    \multirow{7}{*}{\shortstack{Waymo \\ $\downarrow$ \\ KITTI-rain}} & Source-Only & 40.66 & -- & 23.73 & -- & 30.74 & -- & 27.15 & -- & 38.93 & -- & 34.64 & -- \\
    & SN~\cite{wang2020train} & 47.99 & +38.32\% & 30.95 & +30.67\% & 41.19 & +85.80\% & 39.01 & +104.77\% & 36.75 & -16.12\% & 35.82 & +8.03\% \\
    & ST3D~\cite{yang2021st3d} & 57.21 & +86.51\% & 37.57 & +58.79\% & 35.54 & +39.41\% & 34.28 & +62.99\% & 44.54 & +41.49\% & 38.95 & +29.34\% \\
    & UMT~\cite{hegde2021uncertainty} & 55.89 & +79.61\% & 38.99 & +64.83\% & --& --& --& --& --& --& --& --\\
    & ST3D++~\cite{yang2022st3d++} & 59.36 & +97.75\% & 42.27 & +78.76\% & 45.97 & +125.04\% & 43.95 & +148.41\% & 48.63 & +71.75\% & 42.47 & +53.30\% \\
    & \textbf{STAL3D}\cellcolor{LightCyan}  & \textbf{65.85}\cellcolor{LightCyan}  & \textbf{+131.68\%}\cellcolor{LightCyan}  & \textbf{47.60}\cellcolor{LightCyan} & \textbf{+101.40\%}\cellcolor{LightCyan} & \textbf{47.62}\cellcolor{LightCyan}  & \textbf{+138.59\%}\cellcolor{LightCyan}  & \textbf{46.13}\cellcolor{LightCyan} & \textbf{+167.67\%}\cellcolor{LightCyan} & \textbf{53.19}\cellcolor{LightCyan}  & \textbf{+105.47\%}\cellcolor{LightCyan}  & \textbf{46.24}\cellcolor{LightCyan} & \textbf{+78.97\%}\cellcolor{LightCyan}\\
    \cline{2-14}
    & Oracle & 59.79 & +100\% & 47.22 & +100\% & 42.92 & +100\% & 38.47 & +100\% & 52.45 & +100\% & 49.33 & +100\%\\
    \toprule[0.8pt]
    \bottomrule[0.8pt]

\multirow{6}{*}{\shortstack{Lyft \\ $\downarrow$ \\ KITTI-rain}} & Source-Only & 41.27 & -- & 25.63 & -- & 30.26 & -- & 28.11 & -- & 39.45 & -- & 35.28 & -- \\
    & SN~\cite{wang2020train} & 47.53 & +33.80\% & 31.41 & +26.77\% & 40.65 & +82.07\% & 38.59 & +101.16\% & 39.64 & +1.46\% & 35.57 & +2.06\% \\
    & ST3D~\cite{yang2021st3d} & 56.34 & +81.37\% & 39.02 & +62.02\% & 40.84 & +83.57\% & 38.77 & +102.90\% & 45.42 & +45.92\% & 39.33 & +28.83\% \\
    & ST3D++~\cite{yang2022st3d++} & 58.76 & +94.44\% & 41.53 & +73.65\% & 44.85 & +115.24\% & 42.91 & +142.86\% & 48.04 & +66.08\% & 41.80 & +46.41\% \\
    & \textbf{STAL3D}\cellcolor{LightCyan}  & \textbf{65.39}\cellcolor{LightCyan}  & \textbf{+130.24\%}\cellcolor{LightCyan}  & \textbf{47.34}\cellcolor{LightCyan} & \textbf{+100.56\%}\cellcolor{LightCyan} & \textbf{45.93}\cellcolor{LightCyan}  & \textbf{+123.78\%}\cellcolor{LightCyan}  & \textbf{44.32}\cellcolor{LightCyan} & \textbf{+156.47\%}\cellcolor{LightCyan} & \textbf{52.31}\cellcolor{LightCyan}  & \textbf{+98.92\%}\cellcolor{LightCyan}  & \textbf{45.56}\cellcolor{LightCyan} & \textbf{+73.17\%}\cellcolor{LightCyan}\\
    \cline{2-14}
    & Oracle & 59.79 & +100\% & 47.22 & +100\% & 42.92 & +100\% & 38.47 & +100\% & 52.45 & +100\% & 49.33 & +100\%\\
    \toprule[1pt]
    \end{tabular}
    }
    \label{tab2_kitti-rain}
\end{table*}

\begin{figure*}[t]
   \centering
   \includegraphics[width=0.99\linewidth]{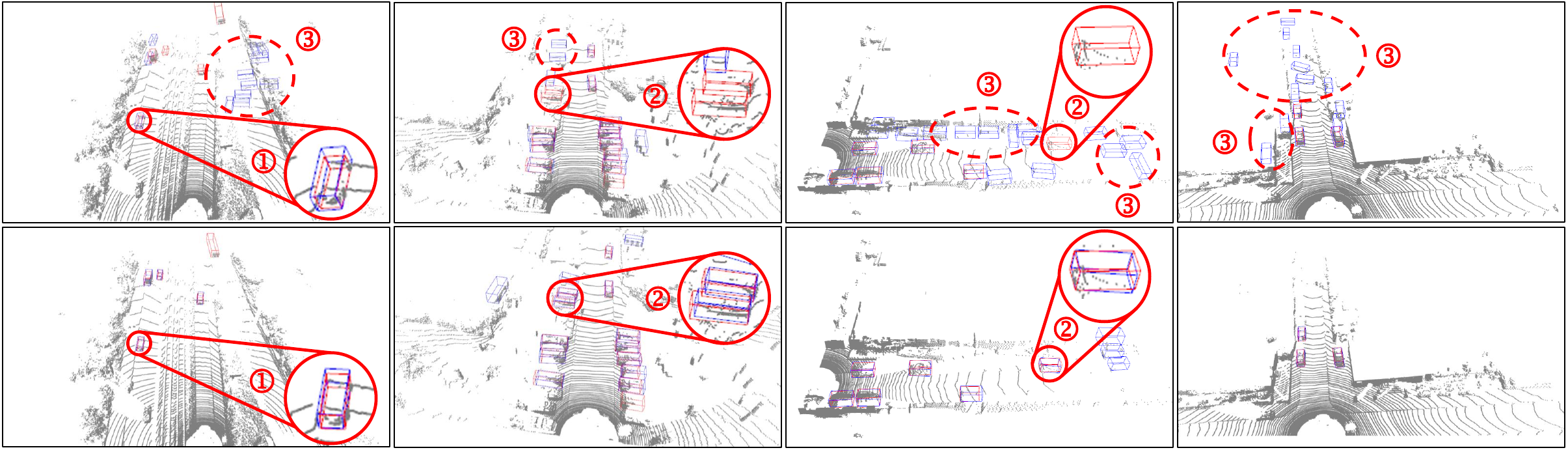}
   \caption{Qualitative results of our proposed STAL3D on nuScenes $\rightarrow$ KITTI, where the top/bottom line represents the pre/post cross-domain detection results, and the blue/red boxes indicate the predicted/GT bounding boxes. \ding{172} alleviates the issue of source domain size bias; \ding{173} reduces missed detections; \ding{174} reduces false positives.}
   \label{visual0}
\end{figure*}

\begin{figure*}[t]
   \centering
   \includegraphics[width=0.99\linewidth]{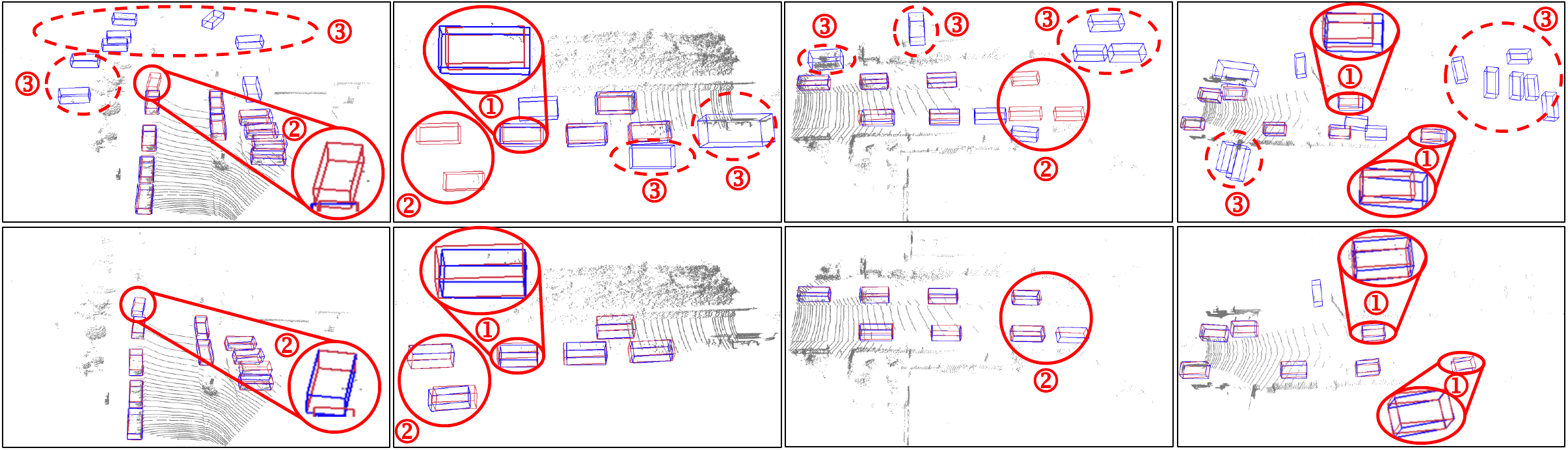}
   \caption{Qualitative results of our proposed STAL3D on Waymo $\rightarrow$ KITTI-rain, where the top/bottom line represents the pre/post cross-domain detection results, and the blue/red boxes indicate the predicted/GT bounding boxes. \ding{172} alleviates the issue of source domain size bias; \ding{173} reduces missed detections; \ding{174} reduces false positives.}
   \label{visual_kitti_rain}
\end{figure*}

\textbf{Evaluation Metric.}
We adhere to the official KITTI metric and present the average precision (AP) in both the bird's-eye view (BEV) and 3D over $40$ recall positions. The mean average precision is assessed with an IoU threshold of $0.7$ for \textit{cars} and $0.5$ for \textit{pedestrians}  and \textit{cyclists}. Additionally, we quantify the closure of the performance gap from Source Only to Oracle, denoted as \textbf{closed gap}
$=\frac{\text{AP}_{\text{model}} - \text{AP}_{\text{source only}}}{\text{AP}_{\text{oracle}} - \text{AP}_{\text{source only}}} \times 100\%.$

\textbf{Implementation Details.} We validate our STAL3D method using SECOND-IoU~\cite{yan2018second}. Pretraining of our detectors on the source domain follows the training settings outlined in the widely-used point cloud detection codebase OpenPCDet \cite{openpcdet2020}. During the subsequent self-training stage on the target domain, we employ Adam with a learning rate of $1.5 \times 10^{-3}$ and utilize a one-cycle scheduler for fine-tuning the detectors over 30 epochs. During the generation of pseudo-labels, the hyperparameter $\varphi$ is set to 0.2. In the Focal loss, $\alpha$ is set to 0.25 and $\gamma$ is set to 2. In the region suppression, $k$ is set to 20\%. For the model optimization objective, $\lambda_{1}$ is set to 1.0, $\lambda_{2}$ is set to 2.0, $\lambda_{3}$ is set to 1.0, $\lambda_{4}$ is set to 0.2, and $\lambda_{5}$ is set to 1.0. All experiments are accomplished on 4 NVIDIA Tesla V100 GPUs.

\subsection{Main Results}

\textbf{Quantitative results.} As shown in Tab.~\ref{tab1_main}, we compare the performance of our STAL3D with Source Only, SN~\cite{wang2020train}, ST3D~\cite{yang2021st3d}, ST3D++~\cite{yang2022st3d++} and Oracle on five adaptation tasks. We can clearly observe that STAL3D consistently improves the performance on Waymo $\rightarrow$ KITTI, Waymo $\rightarrow$ Lyft, Waymo $\rightarrow$ nuScenes, nuScenes $\rightarrow$ KITTI and Lyft $\rightarrow$ KITTI by a large margin of 23.14\%, 5.84\%, 5.01\%, 24.17\% and 15.47\% in terms of mAP$_{3D}$, which largely close the performance gap between Source Only and Oracle. When comparing with the latest SOTA method ST3D++, our STAL3D exhibits superior performance in terms of mAP$_{3D}$ for all five adaptation tasks, with improvements of 3.64\%, 1.42\%, 1.77\%, 1.94\% and 2.58\%, respectively. We attribute these performance improvements to our approach, which unifies self-training and adversarial learning paradigms into one framework, unleashing the potential complementary advantages of pseudo-labels and feature distribution alignment. Additionally, the BS-AL and SFM tailored for 3D cross-domain scenes, effectively alleviating the issues of the large proportion of background interference and source domain size bias.

The domain shift caused by adverse weather is a relatively difficult benchmark in 3D cross-domain settings, as special weather not only affect the data generation mode of LiDAR acquisition equipment, but also generate a large number of noise points. As shown in Tab.~\ref{tab2_kitti-rain}, it is worth noting that the proposed STAL3D gains significantly more performance for the Waymo $\rightarrow$ KITTI-rain and Lyft $\rightarrow$ KITTI-rain (even surpassing Oracle), indicating the STAL3D is more effective for adapting to 3D scenes with larger environmental gaps. Overall, the proposed STAL3D excels all baselines on both mAP$_{3D}$ and mAP$_{BEV}$ across all scenarios of 3D adaptation tasks. We contend that in the presence of significant domain gaps, the self-training paradigm based on pseudo labels, owing to its inherent incapacity to align feature spaces, tends to generate low-quality pseudo labels, thereby falling into the source domain bias trap. Our STAL3D collaborates self-training with adversarial learning, achieving feature distribution alignment between source and target domain, so it can better cope with such challenge.

\begin{table}[t]
\renewcommand\arraystretch{1.15}
    \centering
    \caption{Component ablation on Waymo $\rightarrow$ KITTI.}
    \resizebox{0.90\columnwidth}{!}{
    \begin{tabular}{ccc|cc}\toprule[1pt]
    \multicolumn{3}{c|}{Module} &\multicolumn{2}{c}{SECOND-IoU~\cite{yan2018second}} \\
    ST&BS-AL&SFM & $\text{mAP}_{BEV}$ & $\text{mAP}_{3D}$\\ \midrule
     & & & 78.07 & 54.67  \\
    \checkmark& & & 78.67 & 62.60  \\
    \checkmark&\checkmark& & 78.60 & 65.45  \\
    \checkmark\cellcolor{LightCyan}&\checkmark\cellcolor{LightCyan}&\checkmark\cellcolor{LightCyan} & \textbf{82.26}\cellcolor{LightCyan} & \textbf{69.78}\cellcolor{LightCyan} \\
    \bottomrule
    \end{tabular}
}
    \label{tab3_component}
\end{table}

\textbf{Qualitative results.} We also compare the visual quality of the results against that of the source-only network on nuScenes $\rightarrow$ KITTI and Waymo $\rightarrow$ KITTI-rain. As shown in Fig.~\ref{visual0} and Fig.~\ref{visual_kitti_rain}, our STAL3D can improve the performance of cross-domain 3D object detection from three aspects: (1) alleviates the issue of source domain size bias; (2) reduces missed detections; (3) reduces false positives. The source only trained model frequently suffers from serious false positives. Our STAL3D collaborates the self-training and adversarial learning, aligning feature distribution for large domain gaps through adversarial learning and using self-training to generate high-quality pseudo labels. And a concise and effective scale filtering module is tailored for the source domain size bias. Thanks to the above advantages, STAL3D achieves excellent cross-domain detection results.

\begin{table}[t]
\renewcommand\arraystretch{1.15}
    \centering
    \caption{Ablation to BS-AL on Waymo $\rightarrow$ KITTI.}
    \resizebox{0.65\columnwidth}{!}{
    \begin{tabular}{l|cc}\toprule[1pt]
    \multicolumn{1}{l|}{\multirow{2}*{Setting}} &\multicolumn{2}{c}{SECOND-IoU~\cite{yan2018second}} \\
    ~  & $\text{mAP}_{BEV}$ & $\text{mAP}_{3D}$\\ \midrule
    baseline & 78.58 & 61.93 \\
    \hline
    CA $\beta$=2 & 78.35 & 63.53  \\
    CA $\beta$=4 & 78.46 & 63.74  \\
    \hline
    FRS $k$=0.05 & 78.21 & 64.48  \\
    FRS $k$=0.10 & 78.60 & 65.16  \\
    \textbf{FRS $k$=0.20}\cellcolor{LightCyan} & \textbf{78.60}\cellcolor{LightCyan} & \textbf{65.45}\cellcolor{LightCyan}  \\
    FRS $k$=0.25 & 78.42 & 63.77  \\
    FRS $k$=0.50 & 78.55 & 62.89  \\
    \bottomrule
    \end{tabular}
    }
    \label{tab4_BS-AL}
\end{table}

\subsection{Ablation Study}

\textbf{Effectiveness of each component.} To verify the effectiveness of each module, we first conduct component ablation experiments on the proposed STAL3D. We abbreviate the Self-Training, Adversarial Learning with Background Suppression and Scale Filtering Module as ST, BS-AL and SFM, respectively. As shown in Tab.~\ref{tab3_component}, with Source Only augmented by ROS as the original baseline, applying ST can improve mAP$_{3D}$ by 7.93\%. This indicates that fully utilizing source domain information and performing ST is significantly superior to direct model transfer. Based on this, after adding our BS-AL module, the mAP$_{3D}$ performance is boosted by 2.85\%, and it is further improved by 4.33\% through incorporating our SFM. We attribute this improvement to the feature distribution correction ability of adversarial learning and the further resolution of source domain size bias. Finally, with all these components, the mAP$_{BEV}$ and mAP$_{3D}$ is boosted to 82.26\% and 69.78\% respectively, validating its effectiveness.

\textbf{Ablation to BS-AL.}
To verify the effectiveness of the BS-AL module, we use a naive adversarial learning approach that removes background suppression operations as the baseline. As shown in Tab.~\ref{tab4_BS-AL}, we compare two different methods for extracting and weighting the foreground-background based on Feature Richness Scores (FRS) and Channel Attention (CA), and ablate the hyperparameters. For the CA approach, we obtain the attention map $\mathcal{A}$ by computing the absolute mean of channel values pixel-wise, and then weight the attention map with $\mathcal{A}^{'} = 1 + \beta \cdot \mathcal{A}$. As the results demonstrate, the FRS is the preferable choice, and with the increase of the parameter $k$, the model's detection performance exhibits an initially increasing and subsequently decreasing trend. We postulate that the FRS may not perfectly align with foreground-background. Consequently, with lower $k$, the attention might overlook certain foreground regions, while higher $k$ could result in some interference from background pixels during adversarial learning. Finally, when $k$=0.2, it can bring a performance improvement of 3.52\% on mAP$_{3D}$.

\begin{table}[t]
\renewcommand\arraystretch{1.15}
    \centering
    \caption{Detailed analysis of AL and ST on Waymo $\rightarrow$ KITTI.}
    \resizebox{0.90\columnwidth}{!}{
    \begin{tabular}{l|cc}\toprule[1pt]
    \multicolumn{1}{l|}{\multirow{2}*{Setting}} &\multicolumn{2}{c}{SECOND-IoU~\cite{yan2018second}} \\
    ~ & $\text{mAP}_{BEV}$ & $\text{mAP}_{3D}$\\ \midrule
    Source-Only & 67.64 & 27.48  \\
    + RG\_AL & 49.96 & 17.68  \\
    + F\_AL & 62.83 & 20.19  \\
    \hline
    Source (w/ ROS) & 78.07 & 54.67  \\
    ST (w/ ROS) + Source Label & 76.54 & 52.34 \\
    ST (w/ ROS) & 78.67 & 62.60 \\
    \bottomrule
    \end{tabular}
}
    \label{tab5_AL+ST}
\end{table}

\textbf{Detailed analysis of AL and ST.} As shown in Tab.~\ref{tab5_AL+ST}, we first remove the Self-Training (ST) framework and conducted adversarial training solely using labeled source domain and unlabeled target domain data. Additionally, we devise both feature distribution alignment (F\_AL) and regressor-based distribution alignment (RG\_AL) techniques. RG\_AL refers to the approach introduced by~\cite{jiang2021decoupled}, which employs Generalized Intersection over Union (GIOU)~\cite{rezatofighi2019generalized} as a distance metric for aligning the distribution of bounding boxes in the context of regressors. It can be seen that using these two alignment methods directly cannot improve the results.
This is due to the asymmetrical loss formed as a result of the lack of pseudo-label supervision signals from the target domain in adversarial learning, leading to source domain size bias issue. ROS is an effective data augmentation method that can alleviate source domain size bias to a certain extent, but there is still a 7.93\% gap on mAP$_{3D}$ compared to the ST (w/ ROS). In addition, directly adding Source Label information to ST (w/ROS) can also result in a 10.26\% performance degradation. The results above indicate that both AL and ST are susceptible to the issue of source domain size bias. Additionally, relying solely on ROS is not sufficient for effective resolution.

\begin{table}[t]
\renewcommand\arraystretch{1.15}
    \centering
    \caption{Ablation to SFM on Waymo $\rightarrow$ KITTI.}
    \resizebox{0.90\columnwidth}{!}{
    \begin{tabular}{ll|cc}\toprule[1pt]
    Source Domain & Target Domain  & $\text{mAP}_{BEV}$ & $\text{mAP}_{3D}$\\ \midrule
    cls+reg+IoU & cls+reg+IoU & 78.46 & 63.74  \\
    cls+IoU & cls+reg+IoU & 78.60 & 65.16  \\
    cls+IoU & reg+IoU & 79.24 & 65.67  \\
    \hline
    cls+S\_filter+IoU & cls+S\_filter+IoU & 79.56 & 67.81  \\
    cls+IoU & cls+S\_filter+IoU & 79.31 & 67.58  \\
    \textbf{cls+S\_filter+IoU}\cellcolor{LightCyan} & \textbf{S\_filter+IoU}\cellcolor{LightCyan} & \textbf{79.62}\cellcolor{LightCyan} & \textbf{68.55}\cellcolor{LightCyan}  \\
    \bottomrule
    \end{tabular}
    }
    \label{tab6_SFM}
\end{table}

\begin{figure}[t]
\centering
\includegraphics[width=0.90\columnwidth]{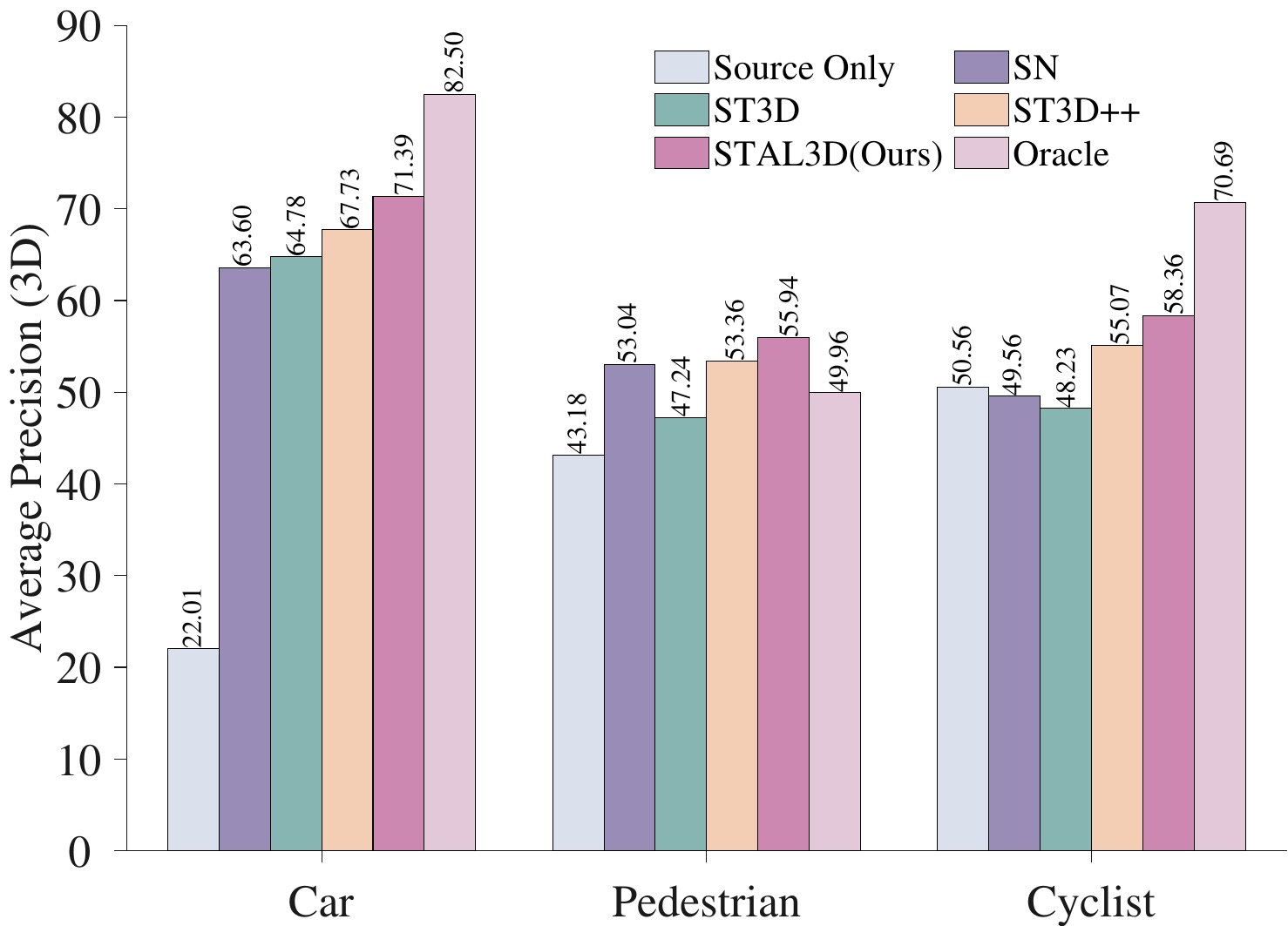}
\caption{Adaptation results based on PV-RCNN. We report $\text{AP}_{3D}$ of Car on Waymo $\rightarrow$ KITTI. }
\label{fig9_results}
\end{figure}

\textbf{Ablation to SFM.} In order to fully validate the effectiveness of SFM, we conducted a detailed combination analysis of the loss term. As shown in Tab.~\ref{tab6_SFM}, without scale filtering, removing the regression supervision signal from the source domain results in a 1.42\% improvement in mAP$_{3D}$. Moreover, comparing mAP$_{3D}$ before and after scale filtering, there is a 4.07\% increase. This demonstrates the necessity of filtering the scale regression term.
Furthermore, it can be observed that regardless of whether scale is filtered in the regression terms, removing the classification loss from the target domain leads to improvements in detection results, with mAP$_{3D}$ increasing by 1.93\% and 0.74\%, respectively. We believe that despite the denoising process during pseudo-label generation, noise is inevitably present. Utilizing accurate class labels from the source domain enables the self-training process to obtain high-quality classification labels. The final combination of the above special designs result in our ultimate SFM module, bringing a 4.81\% performance improvement.

\textbf{Robustness to Detector Architecture.} All previous experiments are conducted on the SECOND-IoU detector. To further validate the robustness of our method across different detectors, we conduct additional experiments on the Waymo $\rightarrow$ KITTI task using the PV-RCNN~\cite{shi2020pv} framework. As shown in Fig.~\ref{fig9_results}, our approach consistently outperforms previous SOTA methods across all three categories, demonstrating its robustness to detector architecture.

\section{Conclusion}
This paper analyzes the strengths and weaknesses of existing 3D unsupervised domain adaptation paradigms
and points out the strong complementarity between  Self-Training (ST) and Adversarial Learning (AL). To unleash the potential advantages of pseudo-labels and feature distribution alignment, a novel cross-domain 3D object detection framework is proposed via collaborating ST and AL, dubbed as STAL3D. Additionally, considering the characteristics of 3d cross-domain scenarios, a Background Suppression Adversarial Learning (BS-AL) module and a Scale Filtering Module (SFM) are proposed,
effectively alleviating the problems of the large proportion of background interference and source domain size bias. Extensive experiments are conducted on multiple cross-domain tasks, and STAL3D reaches a newly state-of-the-art.

\bibliographystyle{IEEEtran}
\bibliography{references.bib}

\end{document}